\documentclass[runningheads]{llncs}
\usepackage[T1]{fontenc}
\providecommand{\text}[1]{\mathrm{#1}}

\usepackage{amssymb}
\usepackage{xcolor}
\usepackage{multirow}
\usepackage{booktabs}
\usepackage{placeins}
\usepackage{wrapfig}

\usepackage{graphicx,verbatim}
\usepackage{amsmath}
\usepackage{subcaption}
\begin{document}
%
\title{Bi-PT: Bidirectional Cross-Attention Point Transformers for Four-Chamber Heart Reconstruction from Sparse Cardiac MRI Data}

\titlerunning{Bi-PT}

\author{
Chenchuhui Hu\inst{1} \and
Shaoming Pan\inst{1} \and
Leon Axel\inst{2} \and
Meng Ye\inst{1}
}

\authorrunning{Hu et al.}

\institute{
Department of Computer Science and Engineering, University of Texas at Arlington, Arlington, TX, USA \\
\email{\{chenchuhui.hu, shaoming.pan, meng.ye\}@uta.edu}
\and
NYU Grossman School of Medicine, New York University, New York, NY, USA \\
\email{leon.axel@nyulangone.org}
}

\maketitle              
\begin{abstract}
We propose \textbf{Bi-PT}, a method for reconstructing 3D four-chamber human heart meshes from clinical sparsely sampled cardiac magnetic resonance imaging (CMR) data. This work addresses the error-prone generation of 3D cardiac shape from a sparse point cloud (SPC) extracted from 2D long-axis and short-axis views used in routine clinical CMR protocols. Bi-PT enables accurate inference of the four-chamber heart mesh from the SPC by learning robust point features via bidirectional point cross-attention between an atlas and the SPC, together with per-point semantic labels that improve correspondence estimation. We formulate the deformation field as a Neural Ordinary Differential Equation (NODE) parameterized by a per-point affine transformation and translation to deform the atlas toward the target heart shape. By learning such a NODE, we can guarantee the deformation field to be a locally affine diffeomorphic deformation. We also integrate a semantic label loss into the Chamfer distance to encourage label-consistent correspondences and add a smoothness regularization to stabilize and improve the learning of the deformation field. Extensive experiments demonstrate that Bi-PT achieves accurate and robust performance compared to baselines. Code is available at \url{https://github.com/Chenchuhui/Bi-PT}.



\end{abstract}
\section{Introduction}
Cardiac magnetic resonance imaging (CMR) is a gold standard for evaluating cardiac morphology and function, and supports a wide range of downstream clinical and research applications. However, CMR acquisition is time-consuming, making dense 3D imaging impractical in routine settings, due to prolonged breath-holding requirement. As a result, standard clinical protocols for CMR often rely on a combination of a few long-axis (LAX) views and a stack of short-axis (SAX) slices to capture the essential cardiac structure. While these views are clinically efficient, they provide only sparse geometric evidence in 3D. In many settings, dense and accurate 3D geometry reconstruction from these sparsely sampled routine views is needed to support image analysis tasks and quantitative measurements, such as mass and volume \cite{history3}, 3D myocardial wall strain computation \cite{history2,haber2000three}, image-guided interventions \cite{suinesiaputra2017statistical}, and biomechanics finite element simulations \cite{wang2013structure}.

3D cardiac shape reconstruction from multi-view sparse 2D CMR has a long history \cite{birecon1,history2,history3}. Traditional approaches, such as statistical shape modeling (SSM) and principal component analysis (PCA), typically start from segmentation masks and solve an iterative optimization problem to fit a template, often a mean shape, to sparse observations~\cite{myronenko2010point}. These methods rely on repeated registration updates and can be computationally expensive; in practice, inference accuracy can degrade when observations are sparse or noisy. More recently, learning-based methods have been explored to offer more accurate and efficient solutions for 2D-to-3D reconstruction under sparse MRI acquisitions \cite{deepreview}. Existing deep learning methods can generally be grouped into four families: point cloud \cite{ma2025heartformer,pc1,ye2023neural}, mesh-based \cite{birecon3,birecon5}, shape-aware \cite{chen2021shape,jayakumar2023sadir}, and volumetric models \cite{he2023dmcvr}. Across these families, diverse architectures including CNN \cite{pc1}, GCN \cite{kong2021deep} and
point transformer~\cite{ye2023neural}
have been adopted, but limitations remain. Specifically, many methods struggle with sparse or irregular input, unstable training, and high computational cost.


In this work, we propose a bidirectional cross-attention Point Transformer (\textbf{Bi-PT}) for reconstructing 3D four-chamber human heart \emph{meshes} from the sparse point cloud (SPC) extracted from routine clinical sparse CMR data. 
Our Bi-PT provides a robust feature learning from an atlas and the SPC to predict the locally affine diffeomorphic deformation between the atlas and the target shape parameterized by a Neural Ordinary Differential Equation (NODE) \cite{chen2018neural}.
We further propose a semantic-aware Chamfer distance loss to ensure label-consistent correspondence learning and integrate a smoothness regularization term into the loss to enable high accuracy of four-chamber heart mesh reconstruction.

Our contributions can be summarized as follows: 
(1) A novel point transformer architecture that captures both local and global information through bidirectional cross-attention between an atlas and the SPC. 
(2) A locally affine diffeomorphic deformation (LADD) model that produces complex and large deformations and ensures topology preserving of the atlas mesh. 
(3) Accurate label-aware 3D reconstruction that incorporates semantic labels into the Chamfer distance loss to encourage label-consistent matching and supports chamber-level accuracy of the reconstructed four chamber heart mesh.
(4) A Laplacian regularization loss to prevent irregular deformation potentially resulting from extra deformation freedom of LADD.

\begin{figure*}[!ht]
    \centering
    \includegraphics[width=\textwidth]{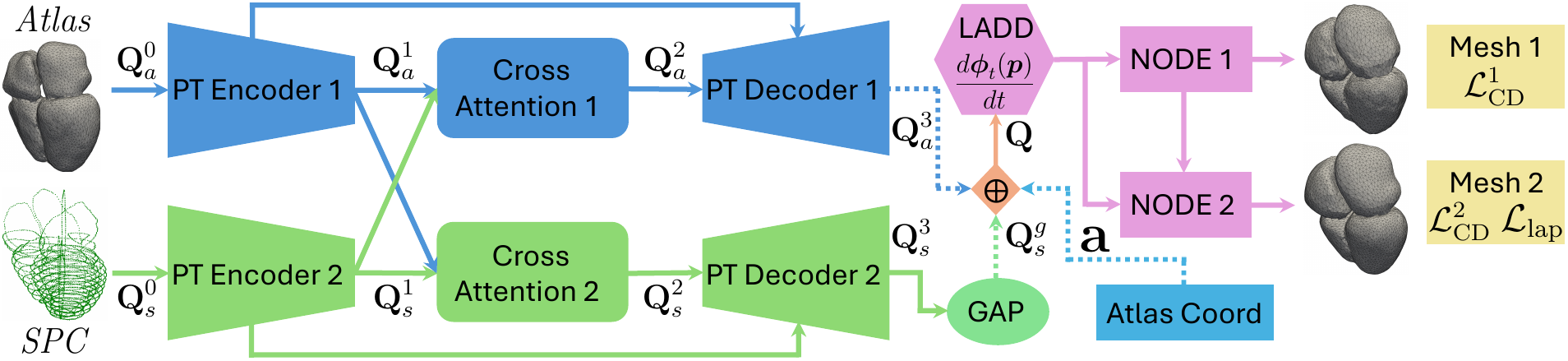}
    \caption{Architecture of \textbf{Bi-PT}. 
    The bidirectional cross-attention Point Transformer (Bi-PT) learns fused atlas-SPC point features, which predict locally affine diffeomorphic deformation (LADD). Two NODE blocks then deform the atlas to produce an intermediate mesh and the final mesh. GAP: global average pooling.
    }
    \label{fig:architecture}
\end{figure*}

\section{Method}

As shown in Fig.~\ref{fig:architecture}, Bi-PT learns an atlas-to-target deformation through four stages: (1) point feature encoding through Point Transformer encoders, (2) bidirectional cross-attention for information fusion, (3) point feature decoding via Point Transformer decoders, and (4) LADD through two NODE blocks. We give details of Bi-PT as follows.

\subsection{Bi-PT Encoder and Decoder}
\label{sec:encoder}
The atlas is shared across all subjects and thus carries no subject-specific evidence; the SPC is the only observation of the individual target. The encoder's role is to give every point of the atlas or SPC a feature summarizing its local neighborhood, so that the bidirectional cross-attention (Sec.~\ref{sec:bca}) can transfer information between subject-specific cues within the SPC and the atlas.

We represent the atlas and the SPC as point sets $\mathbf{Q}_{a}^{0}\in\mathbb{R}^{N_0\times(3+L_a)}$ and $\mathbf{Q}_{s}^{0}\in\mathbb{R}^{M_0\times(3+L_s)}$, each point given by its $3$D
coordinates and a semantic label: a $3$-channel triplet ($L_a=3$) for the atlas and a $2$-channel pair for the SPC ($L_s=2$). See more details in Sec.~\ref{sec:loss}. These label channels make the point features more discriminative than coordinates alone, sharpening the correspondences estimated downstream.

We encode each stream with an encoder of Point Transformer~\cite{zhao2021point}. After a point-wise embedding, its vector self-attention aggregates every point over its $k$ nearest neighbors, with attention weights modulated by a learned position encoding of the neighbors' relative coordinates. 
The point set is progressively coarsened by $B$ transition-down stages, each applying set abstraction~\cite{qi2017pointnet++} 
followed by a self-attention block, doubling the channel width per stage. Applied independently to the two streams, the encoders yields coarse token sets $\mathbf{Q}_{a}^{1}\in\mathbb{R}^{N\times C}$ and $\mathbf{Q}_{s}^{1}\in\mathbb{R}^{M\times C}$ ($N=\frac{1}{256}N_0$, $M=\frac{1}{256}M_0$, $C$ is the channel width).

The coordinate-feature pairs at every level of the encoders are retained as a skip connection for the decoders.
The decoders mirrors the encoders as a $B$-stage upsampling path that propagates the enriched atlas-SPC features from bottleneck tokens back to full resolution.

\subsection{Bidirectional Cross-Attention Core Layer}
\label{sec:bca}
Each encoder processes the atlas or the SPC independently with only self-attention. The resulting tokens, therefore, describe local geometry within each point set but nothing about the correspondence between them. Because the atlas must be deformed onto the shape carried by the SPC, the two token sets must be put in correspondence. We achieve this with our core contribution, a bidirectional cross-attention layer, that exchanges information between the atlas tokens $\mathbf{Q}_{a}^{1}$ and the SPC tokens $\mathbf{Q}_{s}^{1}$ in both directions.

Specifically, for a query token $\mathbf{x}_{q,i}$ and key/value tokens $\mathbf{x}_{k,j}$/$\mathbf{x}_{v,j}$, we linearly project $\mathbf{q}_i = f_q(\mathbf{x}_{q,i})$, $\mathbf{k}_j = f_k(\mathbf{x}_{k,j})$, $\mathbf{v}_j = f_v(\mathbf{x}_{v,j})$, perform a relational encoding $\mathbf{p}_{ij} = \mathrm{MLP}_\delta(\mathbf{x}_{q,i} - \mathbf{x}_{k,j})$, and then compute the \textit{vector} cross-attention~\cite{zhao2021point}: 
\begin{equation}
\alpha_{ij} = \operatorname{softmax}_{j}\!\big(\mathrm{MLP}_\gamma(\mathbf{q}_i-\mathbf{k}_j+\mathbf{p}_{ij})/\sqrt{d}\big),
\label{eq:cross_attn_weight}
\end{equation}
where $f_q$, $f_k$ and $f_v$ are linear layers, $\mathrm{MLP}_\delta$ and $\mathrm{MLP}_\gamma$ are two-layer MLPs with ReLU, and $d$ is the channel dimension. 
The softmax over key index $j$ gives each query a per-key weight distribution. 
Then we aggregate values globally:
\begin{equation}
\mathbf{y}_i = \mathbf{x}_{q,i}+f_o\!\Big(\textstyle\sum_{j=1}^{N_k}\alpha_{ij}\odot(\mathbf{v}_j+\mathbf{p}_{ij})\Big),
\label{eq:cross_attn_aggregation}
\end{equation}
where $f_o$ is a linear projection and $\odot$ is the Hadamard product. 
Crucially, the summation runs over \emph{all} $N_k$ key tokens rather than over a
local neighborhood to ensure global information fusion.

In the $atlas\!\to\!SPC$ direction ($\mathbf{x}_q=\mathbf{Q}_{a}^{1}$,
$\mathbf{x}_k=\mathbf{x}_v=\mathbf{Q}_{s}^{1}$), atlas tokens attend to the SPC to absorb local target-shape cues, yielding $\mathbf{Q}_{a}^{2}\in\mathbb{R}^{N\times C}$; the first decoder  upsamples them to the $N_0$ atlas vertices, producing the per-point features $\mathbf{Q}_{a}^{3}\in\mathbb{R}^{N_0\times C}$.
The reverse $SPC\!\to\!atlas$ direction ($\mathbf{x}_q=\mathbf{Q}_{s}^{1}$, $\mathbf{x}_k=\mathbf{x}_v=\mathbf{Q}_{a}^{1}$) lets SPC tokens aggregate the atlas as a whole, yielding $\mathbf{Q}_{s}^{2}\in\mathbb{R}^{M\times C}$;
the second decoder upsamples them to the $M_0$ sparse points $\mathbf{Q}_{s}^{3}\in\mathbb{R}^{M_0\times C}$, and global average pooling yields a single global target-shape descriptor $\mathbf{Q}_{s}^{g}\in\mathbb{R}^{1\times C}$. We broadcast and concatenate this descriptor to every atlas feature in $\mathbf{Q}_{a}^{3}$. The result together with the atlas coordinates $\mathbf{a}\in\mathbb{R}^{N_0 \times 3}$ give the features $\mathbf{Q}\in\mathbb{R}^{N_0 \times (3 + 2C)}$, which condition the LADD dynamics (Sec.~\ref{sec:ladd}). This design combines local SPC cues with global target-shape context in one module, i.e., the bidirectional cross-attention layer, to improve the robustness of the learned correspondence. 



\subsection{Locally Affine Diffeomorphic Deformation (LADD) Dynamics}
\label{sec:ladd}
The atlas-SPC fusion features $\mathbf{Q}$ predict a time-invariant velocity field parameterized by a NODE which gradually deforms the atlas toward a target heart shape. Unlike per-point translation-only models~\cite{gupta2020neural,ye2023neural}, we introduce a per-point local affine transform $\boldsymbol{A}$ together with a per-point local translation $\boldsymbol{b}$, which increases the expressiveness of the deformation. We define the flow $\boldsymbol{\phi}_t$ by
\begin{equation}
\frac{d\boldsymbol{\phi}_t(\boldsymbol{p})}{dt} = \boldsymbol{A}(\boldsymbol{\phi}_t(\boldsymbol{p}))\,\boldsymbol{\phi}_t(\boldsymbol{p}) + \boldsymbol{b}(\boldsymbol{\phi}_t(\boldsymbol{p})), \quad \boldsymbol{\phi}_0(\boldsymbol{p})=\boldsymbol{p},
\end{equation}
where $\boldsymbol{\phi}_t(\boldsymbol{p})\in\mathbb{R}^3$ denotes the position of an initial point $\boldsymbol{p}$ at time $t$, and $\boldsymbol{A}\in\mathbb{R}^{3\times 3}$ and $\boldsymbol{b}\in\mathbb{R}^{3}$. As in~\cite{gupta2020neural}, we use MLPs to predict $\boldsymbol{A}$ and $\boldsymbol{b}$ from features $\mathbf{Q}$. 
Note that in NDM~\cite{ye2023neural}, large deformation is modeled by global deformation parameter functions. However, it cannot guarantee topology-preserving. Here, the LADD not only enables complex and large deformation, but also preserves topology of the atlas according to the Cauchy-Lipschitz theorem~\cite{brezis2011functional}. 
Similarly as in NDM~\cite{ye2023neural}, we use two consecutive NODE blocks to deform the atlas into the final target heart mesh.

\subsection{Semantic-Aware Chamfer Distance and Laplacian Regularization}
\label{sec:loss}
\paragraph{Semantic-aware Chamfer Distance.}
\begin{wrapfigure}{r}{0.35\textwidth}
  \centering
  \includegraphics[width=0.35\textwidth]{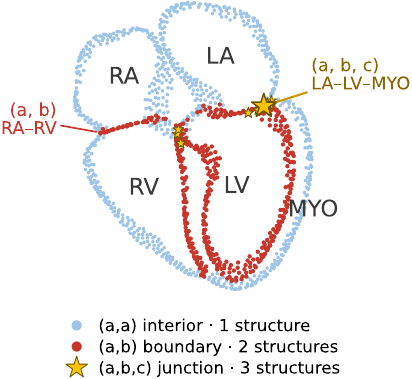}
  \caption{Each point is labeled by the set of chambers it touches:
  $(a,a)$ interior, $(a,b)$ boundary, $(a,b,c)$ junction. 
  }
  \label{fig:sacd}
\end{wrapfigure}
Plain Chamfer distance (CD)~\cite{ye2023neural} across the four chambers ignores chamber identity and can match points across structures. We therefore use a semantic-aware CD (SA-CD) that restricts nearest-neighbor search by the semantic labels. As shown in Fig.~\ref{fig:sacd}, we label each point of the atlas/SPC by the semantics that point has. For a query with boundary-pair label $(a,b)$, $a\neq b$, we first match against points sharing the exact pair $(a,b)$ and, if none exist, relax to points sharing one label (e.g., $(a,c)$ or $(b,c)$); single-structure labels $(a,a)$ keep a strict match. 
The same hierarchy applies to the $3$-label variant (used in Sec.~\ref{sec:encoder} for the atlas). 
This yields label-consistent correspondences (e.g., boundary points match boundary points) while staying robust when a label combination is rare. The reconstruction loss term combines plain CD and semantic-aware CD: $\mathcal{L}_{\mathrm{CD}}=\lambda_{plain}\mathcal{L}_{\mathrm{plain}}+\lambda_{\mathrm{sa}}\mathcal{L}_{\mathrm{sa}}$, where $\lambda_{plain}$ and $\lambda_{\mathrm{sa}}$ are hyper-parameters.
%
We used the $2$-label SA-CD in our default setting, as we found the $3$-label variant could degrade the shape reconstruction accuracy. 

\paragraph{Laplacian regularization.}
Since the LADD dynamics can provide extra freedom on the deformation, we add a Laplacian regularization~\cite{nealen2006laplacian} on the final mesh to promote locally coherent and smooth deformation.

\paragraph{Deep supervision.}
We apply $\mathcal{L}_{\mathrm{CD}}$ to the output of \emph{both} NODE blocks. With $\mathcal{L}_{\mathrm{CD}}^{1}$ and $\mathcal{L}_{\mathrm{CD}}^{2}$ the losses after the first and second block, respectively. The full objective is
\begin{equation}
\mathcal{L}=\lambda_{1}\mathcal{L}_{\mathrm{CD}}^{1}+\lambda_{2}\mathcal{L}_{\mathrm{CD}}^{2}+\lambda_{\mathrm{lap}}\mathcal{L}_{\mathrm{lap}},
\label{eq:total}
\end{equation}
where $\mathcal{L}_{\mathrm{lap}}$ is the Laplacian regularization term and  $\lambda_{1}$, $\lambda_{2}$ and $\lambda_{lap}$ are hyper-parameters.

\section{Experiments}
\subsection{Dataset and Metrics}
\label{ds&metrics}
We evaluated Bi-PT on 1K public cardiac CT cases~\cite{zeng2023imagecas}. We used TotalSegmentator V2~\cite{wasserthal2023totalsegmentator} to segment each case into multiple anatomical structures: right atrium (RA), left atrium (LA), right ventricle (RV), left ventricle (LV), myocardium (MYO), and major vessels. In this work, our shape reconstruction target is the multi-label four-chamber heart mesh (RA/LA/RV/LV/MYO). 
To mimic CMR scans, we standardized each case from the patient coordinate space to clinically meaningful cardiac coordinate space~\cite{xu2024improved} and generated sparse point cloud observations with semantic labels as in~\cite{ye2023neural}, and finally constructed labeled ground-truth meshes from the segmentation masks for supervision and evaluation. 
We performed a subject-level split with 800 cases for training, 100 cases for validation and 100 cases held out for testing. 

For geometric accuracy, we follow prior work~\cite{ye2023neural} and report three point-based distances: (i) Chamfer distance (CD), (ii) Earth Mover's Distance (EMD), and (iii) point-to-surface (P2F) distance. We uniformly sample a dense set of surface points from the predicted mesh and the ground-truth surface, respectively, and compute these metrics between the two sets.
To evaluate mesh quality, we report: (i) normal consistency (NC), which measures agreement of surface normals, (ii) easy non-manifold face (ENF) ratio, which measures the fraction of faces involved in non-manifold configurations, and (iii) self-intersection (SI) ratio, which measures the fraction of faces involved in triangle-triangle intersections. 
Note that, we evaluate the shape reconstruction accuracy of individual structure and report the mean result over the five heart structures. All quantitative evaluations were reported in the original image space.

\subsection{Implementation Details}
\label{sec:implementation}
We implemented Bi-PT in PyTorch and trained all models on NVIDIA A100 GPUs (80\,GB) with a batch size of $5$. We optimized using AdamW (learning rate $5\times10^{-4}$, weight decay $1\times10^{-4}$) for 300 epochs with a cosine annealing schedule, and applied gradient clipping with a maximum norm of $0.1$. 
We set 
$\lambda_{plain}=0.5$,
$\lambda_{sa}=0.5$,
$\lambda_{1}=0.3$, $\lambda_{2}=0.7$
and $\lambda_{\mathrm{lap}}=1.0$ empirically.
We randomly selected a sample heart mesh from the training dataset as the atlas and set $N_0=5632$, $M_0=5632$, $C=512$, $d=512$.
 

\subsection{Baselines}
We compared Bi-PT with one traditional iterative method: Coherent Point Drift (CPD)~\cite{myronenko2010point} and five learning-based baselines: Neural Mesh Flow (NMF)~\cite{gupta2020neural}, MR-Net~\cite{chen2021shape}, Neural Deformable Model (NDM)~\cite{ye2023neural}, Label Transformer Network (LTN) and LTN-DSTN~\cite{xu2024improved}. 
We used the authors' public implementations and matched their recommended training settings as closely as possible.

\begin{figure}[t]
    \centering
    \includegraphics[width=\linewidth]{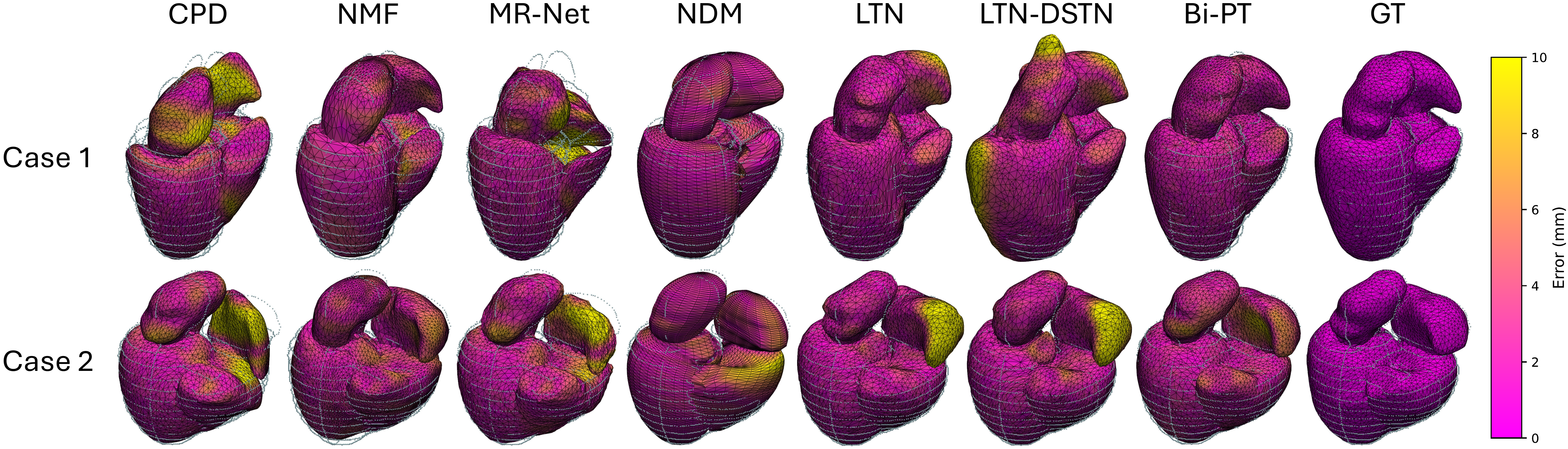}
    \label{fig:baseline}
    \caption{Qualitative comparison of four-chamber whole-heart reconstruction on two representative test cases. For each method, we overlay the predicted mesh on the input SPC; color encodes the per-vertex Euclidean distance to the ground truth, where brighter denotes larger error. 
    }
    \label{fig:baseline}
\end{figure}

\subsection{Results}

\begin{table*}[t]
\centering
\caption{Quantitative comparison with baseline methods on the 100-case test set. CD, EMD, and P2F are reported in \textit{mm}. Values are mean(std) over each chamber. \textcolor{red}{Red} and \textcolor{blue}{blue} indicate the best and second-best results, respectively. 
$\dagger$ indicates a statistically significant improvement over the second-best method using a paired Wilcoxon signed-rank test on the same test cases ($p<0.001$).}
\label{tab:baseline}
\setlength{\tabcolsep}{0.8pt}
\begin{tabular}{l|cccccc}
\toprule
Method & CD~$\downarrow$ & EMD~$\downarrow$ & P2F~$\downarrow$ & NC~$\uparrow$ & ENF~$\downarrow$ & SI($\times10^{-5}$)~$\downarrow$ \\
\midrule
CPD~\cite{myronenko2010point} 
& 5.16(1.96) 
& 12.74(4.99) 
& 4.56(2.06) 
& 0.58(0.05) 
& 0.34(0.00) 
& \textcolor{red}{0(0)} \\

NMF~\cite{gupta2020neural} 
& 2.99(0.48) 
& 13.28(2.49) 
& 2.19(0.56) 
& 0.57(0.04) 
& 0.34(0.00) 
& \textcolor{red}{0(0)} \\

MR-Net~\cite{chen2021shape} 
& 4.90(2.00) 
& 12.21(4.96)
& 4.28(2.10) 
& 0.58(0.07)
& 0.34(0.00) 
& 79.4(355) \\

NDM~\cite{ye2023neural} 
& 3.01(0.47) 
& 12.51(1.61) 
& 2.07(0.55) 
& 0.04(0.03) 
& \textcolor{red}{0.01(0.00)}
& 868.6(967.7) \\

LTN~\cite{xu2024improved} 
& \textcolor{blue}{2.34(0.36)} 
& \textcolor{red}{6.36(1.15)} 
& \textcolor{blue}{1.41(0.40)} 
& \textcolor{blue}{0.61(0.04)}
& 0.35(0.03) 
& \textcolor{blue}{0.56(3.77)} \\

LTN-DSTN~\cite{xu2024improved} 
& 2.62(0.43) 
& 6.96(1.18) 
& 1.77(0.51) 
& 0.60(0.04) 
& 0.36(0.03) 
& 1.17(5.85) \\

\midrule
Bi-PT (Ours) 
& \textcolor{red}{2.28(0.27)}$^\dagger$
& \textcolor{blue}{6.54(1.02)}
& \textcolor{red}{1.41(0.34)}$^\dagger$
& \textcolor{red}{0.66(0.02)}$^\dagger$
& \textcolor{blue}{0.34(0.00)} 
& \textcolor{red}{0(0)}$^\dagger$ \\
\bottomrule
\end{tabular}
\end{table*}

In Tab.~\ref{tab:baseline}, we report the mean quantitative results for four-chamber reconstruction on the test set with the evaluation metrics in Sec.~\ref{ds&metrics}. 
Bi-PT attains the best geometric accuracy while preserving mesh validity. It achieves the lowest CD and P2F distance error, both significant over the second-best method by Wilcoxon test. In terms of NC, it improves a lot compared to the second-best method, indicating a smoother, locally oriented surface. Moreover, Bi-PT produces zero SI, showing that our method retains the atlas-level topology without sacrificing the accuracy. Fig.~\ref{fig:baseline} illustrates examples of the four-chamber reconstruction results. CPD registers the atlas to the SPC purely by minimizing the point-to-point distance, overlaying the atlas as tightly as possible onto the observed points with no awareness of anatomical semantics. 
NMF, being atlas-based, returns overly smooth heart shape reconstruction results with large error in localized regions such as the MYO (LV epi-surface). 
MR-Net behaves almost identical with CPD, with large error patches over both atria and parts of the ventricles. 
NDM uses shape primitives instead of the atlas and cannot accurately recover shape details in the atria and RV.
LTN is the strongest baseline over the bi-ventricular region and the LA but retains visible error on the RA. 
LTN-DSTN performs slightly worse than LTN as it requires stronger smoothness. 
In contrast, Bi-PT stay low-error across all four chambers which we attribute to the bidirectional cross-attention that
injects local cues and global target-shape context of the SPC, the semantic-aware supervision that ties each atlas region to the correct target anatomy, and the Laplacian regularization that keeps smooth shapes.

\subsection{Ablation Studies}
\label{sec:ablation}

From Tab.~\ref{tab:ablation}, for the reconstruction objective, semantic-aware Chamfer consistently improves correspondence-related metrics over plain CD loss (CD 2.28 vs.\ 2.32, EMD 6.54 vs.\ 6.85), indicating that semantic constraints add value beyond geometry-only matching. Bidirectional cross-attention (CA) is critical: replacing bidirectional CA with single CA (sCA), i.e.,  only keeping the $atlas\!\to\!SPC$ pathway without global target-shape descriptor $\mathbf{Q}_{s}^{g}$ injected in $\mathbf{Q}$, leads to a large degradation across metrics (CD 3.10 vs.\ 2.28, EMD 9.04 vs.\ 6.54) and introduces self-intersections, highlighting the importance of the $SPC\!\to\!atlas$
pathway for global target-shape context. For the deformation model, our LADD dynamics improves reconstruction over a per-point translation-only parameterization (CD 2.28 vs.\ 2.32, EMD 6.54 vs.\ 6.56), supporting that the additional degrees of freedom help express complex cardiac shape details. Removing Laplacian regularization increases geometric errors (CD 2.48 vs.\ 2.28, P2F 1.68 vs.\ 1.41) and introduces self-intersections, confirming its role in stabilizing smooth, topology-preserving deformations. Finally, extending the semantics from two to three labels in SA-CD yields no further gain (CD 2.32 vs.\ 2.28), so we adopt the 2-label setting throughout.

\begin{table}[t]
\centering
\caption{Ablation study on Bi-PT. CD, EMD, and P2F are reported in \textit{mm}. Values are mean(std) over each chamber.
S: semantic-aware CD; B: bidirectional cross-attention; 
A: locally affine diffeomorphic deformation;
L: Laplacian regularization. 
Best values are in \textbf{bold}.}
\label{tab:ablation}
\scriptsize
\setlength{\tabcolsep}{2.2pt}
\renewcommand{\arraystretch}{1.05}
\resizebox{\linewidth}{!}{%
\begin{tabular}{@{}l c c c c c c c c c c c@{}}
\toprule
Variant & label & S & B & A & L 
& CD~$\downarrow$ & EMD~$\downarrow$ & P2F~$\downarrow$ 
& NC~$\uparrow$ & ENF~$\downarrow$ & SI($\times10^{-5}$)~$\downarrow$ \\
\midrule
Plain CD 
& 2$l$ & $\times$ & $\checkmark$ & $\checkmark$ & $\checkmark$
& 2.32(0.29) & 6.85(1.15) & 1.46(0.34)
& 0.66(0.02) & 0.34(0.00) & 4.84(26.20) \\

sCA 
& 2$l$ & $\checkmark$ & $\times$ & $\checkmark$ & $\checkmark$
& 3.10(0.62) & 9.04(1.56) & 2.35(0.70)
& 0.60(0.08) & 0.35(0.00) & 2967.20(360.50) \\

Translation
& 2$l$ & $\checkmark$ & $\checkmark$ & $\times$ & $\checkmark$
& 2.32(0.30) & 6.56(1.09) & 1.46(0.36)
& \textbf{0.67(0.02)} & 0.34(0.00) & 0(0) \\

w/o L
& 2$l$ & $\checkmark$ & $\checkmark$ & $\checkmark$ & $\times$
& 2.48(0.36) & \textbf{6.34(0.96)} & 1.68(0.43)
& 0.65(0.02) & 0.34(0.00) & 526.94(242.56) \\

\textbf{Full}
& 2$l$ & $\checkmark$ & $\checkmark$ & $\checkmark$ & $\checkmark$
& \textbf{2.28(0.27)} & 6.54(1.02) & \textbf{1.41(0.34)}
& 0.66(0.02) & \textbf{0.34(0.00)} & \textbf{0(0)} \\
\midrule
Full, 3-label
& 3$l$ & $\checkmark$ & $\checkmark$ & $\checkmark$ & $\checkmark$
& 2.32(0.30) & 7.05(1.17) & 1.46(0.38) & 0.65(0.02) & 0.34(0.00) & 25.60(4.15) \\
\bottomrule
\end{tabular}%
}
\end{table}

\section{Conclusion}
In this work, 
we design Bi-PT, a novel point transformer architecture based on bidirectional cross-attention, for four-chamber heart mesh reconstruction from sparse point cloud observations. 
By combing Bi-PT with locally affine diffeomorphic deformation dynamics, semantic-aware chamfer distance, and Laplacian regularization, our model can ensure accurate reconstruction of patient-specific 3D heart geometry from segmentation boundary of clinical sparse CMR views.

%
%
%
\FloatBarrier
\bibliographystyle{splncs04}
\bibliography{mybibliography}

@article{chen2018neural,
  title={Neural ordinary differential equations},
  author={Chen, Ricky TQ and Rubanova, Yulia and Bettencourt, Jesse and Duvenaud, David K},
  journal={Advances in neural information processing systems},
  volume={31},
  year={2018}
}

@inproceedings{ye2023neural,
  title={Neural deformable models for 3D bi-ventricular heart shape reconstruction and modeling from 2D sparse cardiac magnetic resonance imaging},
  author={Ye, Meng and Yang, Dong and Kanski, Mikael and Axel, Leon and Metaxas, Dimitris},
  booktitle={Proceedings of the IEEE/CVF International Conference on Computer Vision},
  pages={14247--14256},
  year={2023}
}

@article{deepreview,
  title={From 2D to 3D, Deep Learning-based Shape Reconstruction in Magnetic Resonance Imaging: A Review},
  author={McMillian, Emma and Banerjee, Abhirup and Bueno-Orovio, Alfonso},
  journal={arXiv preprint arXiv:2510.01296},
  year={2025}
}

@article{kong2021deep,
  title={A deep-learning approach for direct whole-heart mesh reconstruction},
  author={Kong, Fanwei and Wilson, Nathan and Shadden, Shawn},
  journal={Medical image analysis},
  volume={74},
  pages={102222},
  year={2021},
  publisher={Elsevier}
}

@book{gupta2020neural,
  title={Neural mesh flow: 3d manifold mesh generation via diffeomorphic flows},
  author={Gupta, Kunal},
  year={2020},
  publisher={University of California, San Diego}
}

@article{history2,
  title={Analysis of left ventricular wall motion based on volumetric deformable models and MRI-SPAMM},
  author={Park, Jinah and Metaxas, Dimitris and Axel, Leon},
  year={1996}
}

@article{history3,
  title={Left ventricular mass and volume: fast calculation with guide-point modeling on MR images},
  author={Young, Alistair A and Cowan, Brett R and Thrupp, Steven F and Hedley, Warren J and Dell’Italia, Louis J},
  journal={Radiology},
  volume={216},
  number={2},
  pages={597--602},
  year={2000},
  publisher={Radiological Society of North America}
}

@article{pc1,
  title={Multi-class point cloud completion networks for 3D cardiac anatomy reconstruction from cine magnetic resonance images},
  author={Beetz, Marcel and Banerjee, Abhirup and Ossenberg-Engels, Julius and Grau, Vicente},
  journal={Medical image analysis},
  volume={90},
  pages={102975},
  year={2023},
  publisher={Elsevier}
}

@article{birecon1,
  title={A completely automated pipeline for 3D reconstruction of human heart from 2D cine magnetic resonance slices},
  author={Banerjee, Abhirup and Camps, Juli{\`a} and Zacur, Ernesto and Andrews, Christopher M and Rudy, Yoram and Choudhury, Robin P and Rodriguez, Blanca and Grau, Vicente},
  journal={Philosophical Transactions of the Royal Society A: Mathematical, Physical and Engineering Sciences},
  volume={379},
  number={2212},
  year={2021},
  publisher={The Royal Society}
}

@inproceedings{zhao2021point,
  title={Point transformer},
  author={Zhao, Hengshuang and Jiang, Li and Jia, Jiaya and Torr, Philip HS and Koltun, Vladlen},
  booktitle={Proceedings of the IEEE/CVF international conference on computer vision},
  pages={16259--16268},
  year={2021}
}

@inproceedings{birecon3,
  title={Point2Mesh-Net: Combining point cloud and mesh-based deep learning for cardiac shape reconstruction},
  author={Beetz, Marcel and Banerjee, Abhirup and Grau, Vicente},
  booktitle={International Workshop on Statistical Atlases and Computational Models of the Heart},
  pages={280--290},
  year={2022},
  organization={Springer}
}

@article{birecon5,
  title={Topology-Preserving Loss for Accurate and Anatomically Consistent Cardiac Mesh Reconstruction},
  author={Zhang, Chenyu and Luo, Yihao and Wu, Yinzhe and Hwai Yap, Choon and Yang, Guang},
  journal={arXiv e-prints},
  pages={arXiv--2503},
  year={2025}
}

@article{ma2025heartformer,
  title={HeartFormer: Semantic-Aware Dual-Structure Transformers for 3D Four-Chamber Cardiac Point Cloud Reconstruction},
  author={Ma, Zhengda and Banerjee, Abhirup},
  journal={arXiv preprint arXiv:2512.00264},
  year={2025}
}

@inproceedings{he2023dmcvr,
  title={Dmcvr: Morphology-guided diffusion model for 3d cardiac volume reconstruction},
  author={He, Xiaoxiao and Tan, Chaowei and Han, Ligong and Liu, Bo and Axel, Leon and Li, Kang and Metaxas, Dimitris N},
  booktitle={International conference on medical image computing and computer-assisted intervention},
  pages={132--142},
  year={2023},
  organization={Springer}
}

@article{haber2000three,
  title={Three-dimensional motion reconstruction and analysis of the right ventricle using tagged MRI},
  author={Haber, Idith and Metaxas, Dimitris N and Axel, Leon},
  journal={Medical image analysis},
  volume={4},
  number={4},
  pages={335--355},
  year={2000},
  publisher={Elsevier}
}

@article{suinesiaputra2017statistical,
  title={Statistical shape modeling of the left ventricle: myocardial infarct classification challenge},
  author={Suinesiaputra, Avan and Ablin, Pierre and Alba, Xenia and Alessandrini, Martino and Allen, Jack and Bai, Wenjia and Cimen, Serkan and Claes, Peter and Cowan, Brett R and D’hooge, Jan and others},
  journal={IEEE journal of biomedical and health informatics},
  volume={22},
  number={2},
  pages={503--515},
  year={2017},
  publisher={IEEE}
}

@article{wang2013structure,
  title={Structure-based finite strain modelling of the human left ventricle in diastole},
  author={Wang, HM and Gao, H and Luo, XY and Berry, C and Griffith, BE and Ogden, RW and Wang, TJ},
  journal={International journal for numerical methods in biomedical engineering},
  volume={29},
  number={1},
  pages={83--103},
  year={2013},
  publisher={Wiley Online Library}
}

@inproceedings{jayakumar2023sadir,
  title={SADIR: shape-aware diffusion models for 3D image reconstruction},
  author={Jayakumar, Nivetha and Hossain, Tonmoy and Zhang, Miaomiao},
  booktitle={International workshop on shape in medical imaging},
  pages={287--300},
  year={2023},
  organization={Springer}
}

@article{zeng2023imagecas,
  title={ImageCAS: A large-scale dataset and benchmark for coronary artery segmentation based on computed tomography angiography images},
  author={Zeng, An and Wu, Chunbiao and Lin, Guisen and Xie, Wen and Hong, Jin and Huang, Meiping and Zhuang, Jian and Bi, Shanshan and Pan, Dan and Ullah, Najeeb and others},
  journal={Computerized Medical Imaging and Graphics},
  volume={109},
  pages={102287},
  year={2023},
  publisher={Elsevier}
}

@article{wasserthal2023totalsegmentator,
  title={TotalSegmentator: robust segmentation of 104 anatomic structures in CT images},
  author={Wasserthal, Jakob and Breit, Hanns-Christian and Meyer, Manfred T and Pradella, Maurice and Hinck, Daniel and Sauter, Alexander W and Heye, Tobias and Boll, Daniel T and Cyriac, Joshy and Yang, Shan and others},
  journal={Radiology: Artificial Intelligence},
  volume={5},
  number={5},
  pages={e230024},
  year={2023},
  publisher={Radiological Society of North America}
}

@inproceedings{xu2024improved,
  title={Improved 3D whole heart geometry from sparse CMR slices},
  author={Xu, Yiyang and Xu, Hao and Sinclair, Matthew and Puyol-Ant{\'o}n, Esther and Niederer, Steven A and Chiribiri, Amedeo and Williams, Steven E and Williams, Michelle C and Young, Alistair A},
  booktitle={International Workshop on Statistical Atlases and Computational Models of the Heart},
  pages={43--52},
  year={2024},
  organization={Springer}
}

@book{brezis2011functional,
  title={Functional analysis, Sobolev spaces and partial differential equations},
  author={Brezis, Haim and Br{\'e}zis, Haim},
  volume={2},
  number={3},
  year={2011},
  publisher={Springer}
}

@inproceedings{nealen2006laplacian,
  title={Laplacian mesh optimization},
  author={Nealen, Andrew and Igarashi, Takeo and Sorkine, Olga and Alexa, Marc},
  booktitle={Proceedings of the 4th international conference on Computer graphics and interactive techniques in Australasia and Southeast Asia},
  pages={381--389},
  year={2006}
}

@article{qi2017pointnet++,
  title={Pointnet++: Deep hierarchical feature learning on point sets in a metric space},
  author={Qi, Charles Ruizhongtai and Yi, Li and Su, Hao and Guibas, Leonidas J},
  journal={Advances in neural information processing systems},
  volume={30},
  year={2017}
}

@article{myronenko2010point,
  title={Point set registration: Coherent point drift},
  author={Myronenko, Andriy and Song, Xubo},
  journal={IEEE transactions on pattern analysis and machine intelligence},
  volume={32},
  number={12},
  pages={2262--2275},
  year={2010},
  publisher={IEEE}
}

@article{chen2021shape,
  title={Shape registration with learned deformations for 3D shape reconstruction from sparse and incomplete point clouds},
  author={Chen, Xiang and Ravikumar, Nishant and Xia, Yan and Attar, Rahman and Diaz-Pinto, Andres and Piechnik, Stefan K and Neubauer, Stefan and Petersen, Steffen E and Frangi, Alejandro F},
  journal={Medical Image Analysis},
  volume={74},
  pages={102228},
  year={2021},
  publisher={Elsevier}
}





\end{document}